\documentclass[11pt]{article}
\usepackage[a4paper,margin=1in]{geometry}
\usepackage{titlesec}
\titlespacing*{\section}{0pt}{12pt plus 4pt minus 2pt}{6pt plus 2pt minus 2pt}
\titlespacing*{\subsection}{0pt}{10pt plus 3pt minus 2pt}{4pt plus 2pt minus 2pt}
\titlespacing*{\subsubsection}{0pt}{8pt plus 2pt minus 2pt}{3pt plus 2pt minus 2pt}
\titlespacing*{\paragraph}{0pt}{6pt plus 2pt minus 1pt}{1em}
\usepackage{libertinus} 
\usepackage{microtype}
\usepackage{abstract}
\usepackage{url}
\usepackage[colorlinks=true,linkcolor=black,citecolor=blue!50!black,urlcolor=blue!50!black]{hyperref}
\usepackage{graphicx}
\usepackage{booktabs}
\usepackage{tabularx}

\usepackage[font=small,labelfont=bf,skip=8pt]{caption}
\usepackage{seqsplit}
\newcolumntype{Y}{>{\raggedright\arraybackslash}X}
\usepackage{enumitem}
\usepackage{xcolor}
\usepackage{listings}
\usepackage[scaled=0.9]{inconsolata}
\usepackage[numbers,sort&compress]{natbib}
\usepackage{fancyhdr}
\fancypagestyle{plain}{%
  \fancyhf{}%
  \fancyhead[R]{\small\textit{DOI:} \href{https://doi.org/10.5281/zenodo.17464706}{\texttt{10.5281/zenodo.17464706}}}%
  \fancyfoot[C]{\thepage}%
}
\setlist{
    topsep=6pt,        
    itemsep=3pt,       
    parsep=0pt,        
    partopsep=0pt      
}

\definecolor{jsonkeys}{RGB}{0,0,139}        
\definecolor{jsonstrings}{RGB}{139,0,0}     
\definecolor{jsonnumbers}{RGB}{0,100,0}     
\definecolor{jsonkeywords}{RGB}{75,75,150}  
\definecolor{jsoncomments}{RGB}{128,128,128} 
\definecolor{jsonpunctuation}{RGB}{64,64,64} 

\lstdefinelanguage{json}{
  basicstyle=\ttfamily\small,
  numbers=left,
  numberstyle=\tiny\color{gray},
  numbersep=8pt,
  showstringspaces=false,
  breaklines=true,
  frame=lines,
  backgroundcolor=\color{white},
  keywords={true,false,null},
  keywordstyle=\color{jsonkeywords},
  literate=
   *{:}{{{\color{jsonpunctuation}{:}}}}{1}
    {,}{{{\color{jsonpunctuation}{,}}}}{1}
    {\{}{{{\color{jsonpunctuation}{\{}}}}{1}
    {\}}{{{\color{jsonpunctuation}{\}}}}}{1}
    {[}{{{\color{jsonpunctuation}{[}}}}{1}
    {]}{{{\color{jsonpunctuation}{]}}}}{1}
    {0}{{{\color{jsonnumbers}{0}}}}{1}
    {1}{{{\color{jsonnumbers}{1}}}}{1}
    {2}{{{\color{jsonnumbers}{2}}}}{1}
    {3}{{{\color{jsonnumbers}{3}}}}{1}
    {4}{{{\color{jsonnumbers}{4}}}}{1}
    {5}{{{\color{jsonnumbers}{5}}}}{1}
    {6}{{{\color{jsonnumbers}{6}}}}{1}
    {7}{{{\color{jsonnumbers}{7}}}}{1}
    {8}{{{\color{jsonnumbers}{8}}}}{1}
    {9}{{{\color{jsonnumbers}{9}}}}{1}
    {"}{{{\color{jsonkeys}{"}}}}1,
  string=[s]{"}{"},
  stringstyle=\color{jsonstrings},
  comment=[l]{//},
  commentstyle=\color{jsoncomments}\ttfamily,
}

\title{Policy Cards: Machine-Readable Runtime Governance for Autonomous AI Agents}
\author{Juraj Mavračić \\ Symbiotic Dynamics Ltd}
\date{19 October 2025\vspace{-1ex}}

\begin{document}
\maketitle

\begin{abstract}
Policy Cards are introduced as a machine-readable, deployment-layer standard for expressing operational, regulatory, and ethical constraints for AI agents. The Policy Card sits with the agent and enables it to follow required constraints at runtime. It tells the agent what it must and must not do. As such, it becomes an integral part of the deployed agent. Policy Cards extend existing transparency artifacts such as Model, Data, and System Cards by defining a normative layer that encodes allow/deny rules, obligations, evidentiary requirements, and crosswalk mappings to assurance frameworks including NIST AI RMF, ISO/IEC 42001, and the EU AI Act. Each Policy Card can be validated automatically, version-controlled, and linked to runtime enforcement or continuous-audit pipelines. The framework enables verifiable compliance for autonomous agents, forming a foundation for distributed assurance in multi-agent ecosystems. Policy Cards provide a practical mechanism for integrating high-level governance with hands-on engineering practice and enabling accountable autonomy at scale.
\end{abstract}

\section{Introduction}

Deployed AI systems increasingly act as autonomous agents that perceive, reason, and execute actions through APIs, sensors, and physical actuators in domains constrained by law, ethics, and safety-critical procedures. As these systems move from controlled development settings to operational environments, they must adhere to binding rules governing what actions are permissible, under which conditions exceptions apply, and how conformance is evidenced. Existing transparency instruments, such as \emph{Model Cards} \citep{mitchell_model_2019}, \emph{Data Cards} \citep{pushkarna_data_2022}, and \emph{System Cards} \citep{openai_gpt-4_2023}, focus on documenting capabilities, limitations, and provenance. They do not provide a formal, auditable specification of operational policy at deployment time, for the agent and human to use.

We introduce \emph{Policy Cards}, a deployment-layer, normative, and audit-oriented specification for AI systems and agents. A Policy Card encodes the concrete operational constraints of a deployed system. It includes allowed and denied actions, escalation requirements, time-bound exceptions, evidentiary logging, and mapping to governance frameworks in a structured, machine-readable format based on JSON Schema (2020-12). Each card can be validated automatically, diffed between versions, and integrated with runtime enforcement or testing pipelines. Policy Cards extend the family of transparency artifacts by defining the binding layer of AI assurance.

\begin{figure}[h]
\centering
\includegraphics[width=0.33\linewidth]{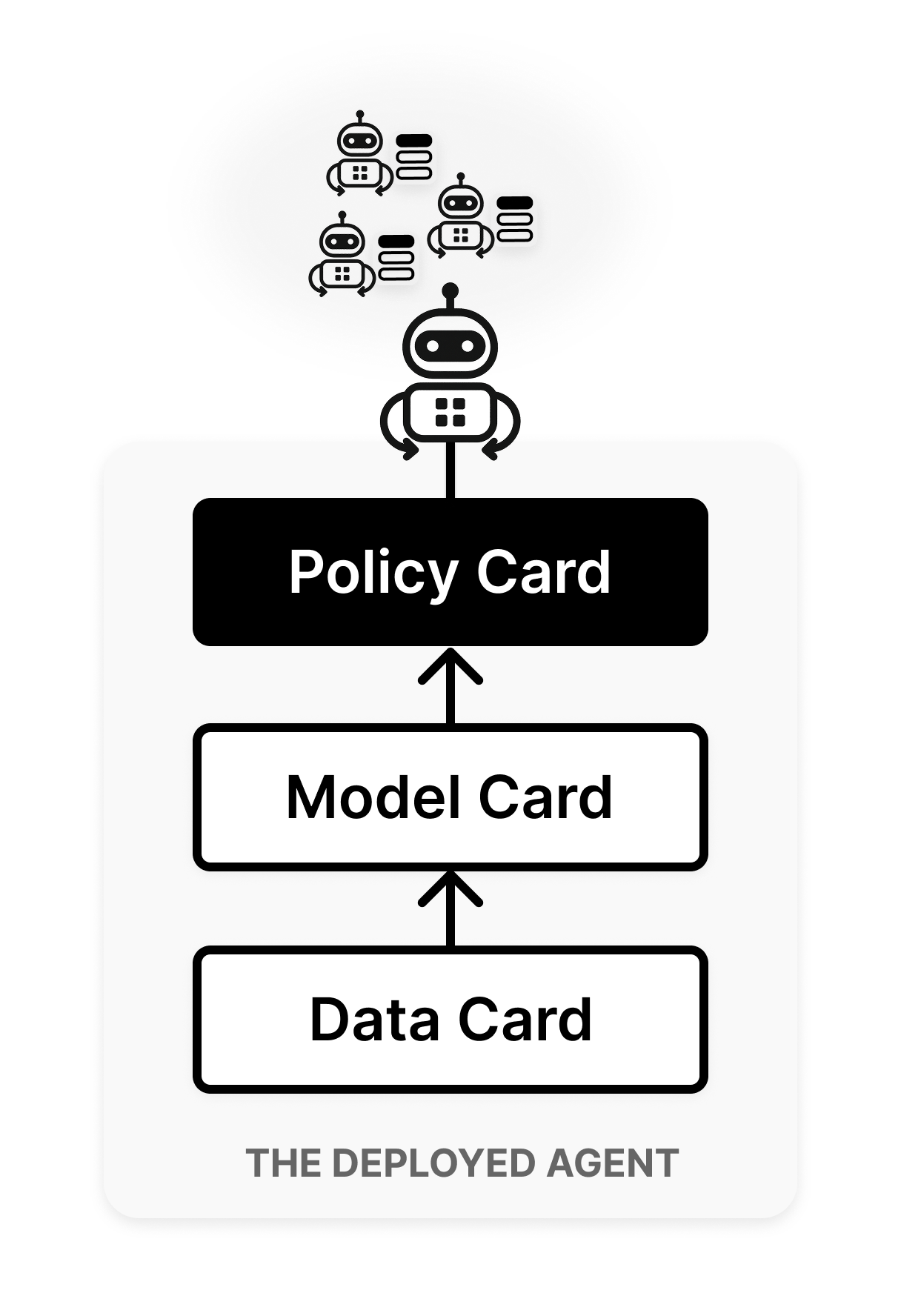}
\caption{Policy cards sit with the deployed agent and tell it what is must and must not do. Agents governed by different Policy Cards behave differently.}
\label{fig:policy-cards}
\end{figure}

As such, Policy Cards benefit both humans and machines. The Policy Card sits with an agent, guiding it on what it must and must not do (Figure \ref{fig:policy-cards}). It defines the agent's operational constraints explicitly. Since two deployed agents based on the same model but using different Policy Cards will behave differently, the Policy Card becomes an integral part of the deployed agent. This is analogous to a human individual constrained by their role within an organization. The agent cannot be expected to behave in a compliant manner without clarity about what those expectations are. Mutual trust can only be established when the expectations are mutually understood.

In practice, organizations distribute obligations across heterogeneous sources, such as legal contracts, risk registers, and process documentation. When AI agents are added to the organizational ecosystem, the absence of a machine-readable policy layer leads to inconsistent governance and difficulty in demonstrating compliance to regulators. The dispersion of obligations across many sources obstructs automated validation, introduces policy drift, and makes audits reactive instead of continuous. What is missing is a unifying artifact capable of connecting engineering reality and regulatory expectation. Such an artifact needs to be versioned, verified, and linked to the operational runtime of an AI system.

\paragraph{Contributions.}This work makes the following contributions:
\begin{enumerate}
\item \emph{Definition and scope.} We formalize Policy Cards as a machine-readable, version-controlled artifact expressing operational rules, obligations, and evidence requirements tied to a specific deployment context and jurisdiction.
\item \emph{Schema and validator.} We present a concrete JSON Schema that enforces syntactic and semantic constraints, including temporal bounds, versioning, and critical auto-fail conditions, and a validator with lint rules for consistency and completeness.
\item \emph{Integration framework.} We describe how Policy Cards integrate with \emph{Declare-Do-Audit} workflows, supporting continuous assurance, CI/CD gating, and stress-testing of deployed agents.
\item \emph{Regulatory crosswalk.} We map schema sections to key assurance frameworks: NIST AI RMF, ISO/IEC 42001, and EU AI Act Annex IV + Article 72, to demonstrate interoperability and compliance readiness.
\item \emph{Domain exemplars.} We provide validated examples in finance, healthcare, and defence domains, illustrating how a single schema accommodates domain-specific regulatory and evidentiary needs.
\item \emph{Forward-looking extensions.} We outline how Policy Cards generalize toward agent-readable governance, multi-agent assurance networks, and cryptographic verification mechanisms.
\end{enumerate}

\paragraph{Positioning.}Policy Cards complement existing AI-agentic transparency tools by introducing a normative governance artifact directly tied to deployment. Whereas Model, Data, and System Cards describe how AI systems are built and evaluated, Policy Cards specify how they must operate under real-world constraints. They enable organizations, auditors, and regulators to verify compliance at design and runtime, instead of retrospectively. Their machine-readable structure allows both humans and agents to interpret and enforce operational norms, making Policy Cards a foundational component for scalable, verifiable AI governance.


\section{Background \& Related Work}

\subsection{Transparency Artifacts}

A growing family of transparency documentation instruments aims to make AI systems more interpretable and accountable. \emph{Model Cards} \citep{mitchell_model_2019} describe trained models, detailing their intended use, evaluation datasets, performance metrics, and their known limitations. \emph{Data Cards} \citep{pushkarna_data_2022} capture training-data provenance, composition, consent status, and known biases. \emph{System Cards} \citep{openai_gpt-4_2023} summarize risks, mitigations, and performance metrics at the system level for deployed models. These artifacts improve transparency across the model lifecycle and have been incorporated into emerging industry practice and guidance from organizations such as Google, OpenAI, and the Partnership on AI.

Despite their value, these documentation artifacts remain largely descriptive. They communicate how an AI system was trained or tested, but not how it must behave under operational, regulatory, or contractual constraints. None of them formalizes \emph{binding deployment policy}, i.e., the specific rules and evidentiary requirements that govern an AI system in situ (Table \ref{tab:positioning}). Consequently, organizations lack a standardized mechanism to specify conditions of use, escalation requirements, exception handling, or audit criteria in a machine-readable format.
It is noted that an AI agent can only be assured if its governing requirements are considered part of the deployed model itself, since the behaviour of an agent built on the same underlying model can vary depending on the operational constraints.

Policy Cards extend the existing ecosystem by addressing the missing normative layer. Instead of reporting model characteristics or data provenance, they define the operational logic that a deployed system must obey and the evidence that must accompany its decisions. This distinction moves AI documentation from retrospective transparency toward agentically proactive and auditable governance.

\subsection{Assurance and Governance Frameworks}

Several international standards and policy frameworks define what should be controlled or documented when deploying AI systems, but they stop short of prescribing a concrete technical artifact. The \emph{NIST AI Risk Management Framework (AI RMF)} \citep{nist} structures responsible AI governance through four functions: Govern, Map, Measure, and Manage. It emphasizes documentation of context, measurement, and monitoring. The \emph{ISO/IEC 42001:2023} \citep{iso_IEC} standard introduces an \emph{AI Management System (AIMS)} model, analogous to ISO 9001 or ISO 27001, requiring procedures for context analysis, operational planning, and continuous performance evaluation. The \emph{EU AI Act (2024)} \citep{european_parliament_and_council_regulation_2024} specifies obligations for high-risk systems, including comprehensive technical documentation (Annex IV) and post-market monitoring requirements (Article 72). These frameworks share the goal of traceability, accountability, and risk control across the AI lifecycle.

However, all three frameworks are primarily textual or process-oriented. They define what information must exist but not how to represent it in a verifiable, machine-readable structure. As a result, compliance remains reliant on human interpretation, fragmented evidence, and manual audits. Policy Cards operationalize these governance expectations through a unified schema that explicitly links every rule, control, and assurance requirement to measurable evidence. Each schema section can be traced to clauses or functions within these frameworks, enabling automated validation of compliance claims.

\subsection{Practical Need for a Deployment-Layer Artifact}

Policy Cards respond to a very practical need. AI systems increasingly execute actions that have legal and ethical consequences. Banks, hospitals, and public agencies require guarantees that decisions made by algorithmic systems comply with sectoral laws, security controls, and safety standards. Traditional documentation cannot enforce or verify such guarantees. Without a deployment-layer artifact, rules are scattered across policy documents, system configurations, and code annotations, which are all prone to drift and ambiguity.

A Policy Card solves this problem by consolidating obligations, controls, and evidence requirements into a single, verifiable schema that can be referenced in audits, CI pipelines, or runtime enforcement. It serves as a contract between the system and its oversight mechanisms, ensuring that policy expectations are explicit, testable, and versioned. This unification of technical and regulatory semantics represents a necessary evolution from descriptive transparency to enforceable assurance.

\begin{table}[h]
\centering
\small
\begin{tabularx}{\textwidth}{@{}lYYY@{}}
\toprule
Artifact & Primary Scope & Normativity & Machine-Readable Audit Hooks \\
\midrule
Model Card & Trained model reporting & Descriptive & Limited (links/IDs) \\
Data Card & Dataset provenance & Descriptive & Limited (links/IDs) \\
System Card & System-level risks/mitigations & Descriptive/Advisory & Limited (summary) \\
	Policy Card (this work) & Deployment policy for a specific system/context & Normative (allow/deny/escalate) & Yes (rules, evidence, thresholds, mappings) \\
\bottomrule
\end{tabularx}
\caption{Positioning: Policy Cards complement existing artifacts with a normative, deployment-bound specification and explicit audit hooks.}
\label{tab:positioning}
\end{table}
\section{Policy Cards: Definition \& Design Goals}

\subsection{Definition}

A \emph{Policy Card} is a versioned, machine-readable specification that defines the operational rules, obligations, exceptions, evidentiary requirements, and assurance mappings that govern a deployed AI system or agent within a defined application, stakeholder set, and jurisdiction. Each card encodes what a system must do and how compliance is demonstrated and verified. It is intended to be both human and machine readable, serving as the authoritative reference for runtime enforcement, auditing, and continuous assurance.

Formally, a Policy Card instance conforms to a JSON Schema (2020-12) definition that enforces type safety, pattern constraints, and required properties. The schema specifies normative structures such as \texttt{action\_rules}, \texttt{exceptions}, \texttt{obligations}, and \texttt{monitoring} blocks, each of which can be validated automatically. Each instance carries a unique semantic version, timestamps, and contextual metadata (system ID, jurisdiction, assurance owner) ensuring traceability over the lifecycle of the AI system. By design, a Policy Card can be processed by validators, linked to CI pipelines, or consumed directly by runtime agents.

\subsection{Design Goals}

Policy Cards were developed to meet a set of design principles derived from both engineering practice and regulatory requirements:

\begin{enumerate}
    \item \emph{Binding and Auditable.} Policy Cards express normative constraints in a verifiable format. Each rule is explicit about its effect (\texttt{allow}, \texttt{deny}, or \texttt{require\_escalation}), and ties directly to measurable evidence fields, critical thresholds, and retention cadences. This enables deterministic compliance checking and facilitates continuous audit trails rather than periodic manual reviews.
    \item \emph{Contextualized.} Each card is scoped to a specific deployment context: domain, jurisdiction, data sensitivity level, and operational boundaries. Contextualization ensures that obligations and controls reflect the applicable legal, ethical, and organizational frameworks, preventing generic or ambiguous statements that cannot be enforced.
    \item \emph{Machine-Readable and Diff-Able.} Policy Cards are structured as JSON documents, compatible with version control systems and automated comparison tools. Semantic versioning (\texttt{X.Y.Z}) enables traceable updates, while diff-based analysis allows auditors to verify what changed between policy revisions. This design principle supports configuration management, regulatory filings, and reproducible assurance.
    \item \emph{Composable.} Policy Cards can reference related artifacts such as Model, Data, or System Cards, Data Protection Impact Assessments (DPIAs), or risk registers. Through unique identifiers and URIs, dependencies remain transparent and maintainable. This composability allows modular governance, where policies for subcomponents can be assembled into system-level composites.
    \item \emph{Actionable.} Every entry in a Policy Card can be operationalized by enforcement engines, monitoring subsystems, or testing frameworks. The schema is compatible with access-control logics such as ABAC (Attribute-Based Access Control) and with structured logging formats for audit evidence. This enables direct integration with CI/CD pipelines, runtime gateways, or automated test harnesses.
    \item \emph{Extensible and Future-Proof.} The schema provides optional ~ fields ~ for ~~\texttt{assurance\_mapping} and \texttt{references}, allowing alignment with new or domain-specific standards without structural changes. Future extensions can include cryptographic attestations or agent-level verification protocols while maintaining backward compatibility.
\end{enumerate}

\subsection{Threat Model and Failure Modes}

Policy Cards are designed to mitigate key governance risks observed in operational AI systems:

\begin{itemize}
    \item \emph{Policy Violations:} Unintended or unsafe actions due to missing or inconsistent constraints. Cards make such policies explicit and enforceable.
    \item \emph{Policy Drift:} Divergence between documented obligations and deployed configurations. Semantic versioning and automated validation prevent silent drift.
    \item \emph{Audit Gaps:} Lack of sufficient evidence or inconsistent assurance records. Defined logging fields and evidence mappings guarantee traceability.
    \item \emph{Unverifiable Claims:} Regulatory submissions without technical grounding. Policy Cards provide structured proof of controls and references to independent standards.
\end{itemize}
Through these mechanisms, Policy Cards establish a linkage between governance requirements and executable verification, establishing a new baseline for deployment-layer assurance in AI systems.

\section{Schema Overview and Validation Tooling} \label{sec:schema}

\subsection{Schema Architecture}

\begin{figure}[h]
\centering
\includegraphics[width=\linewidth]{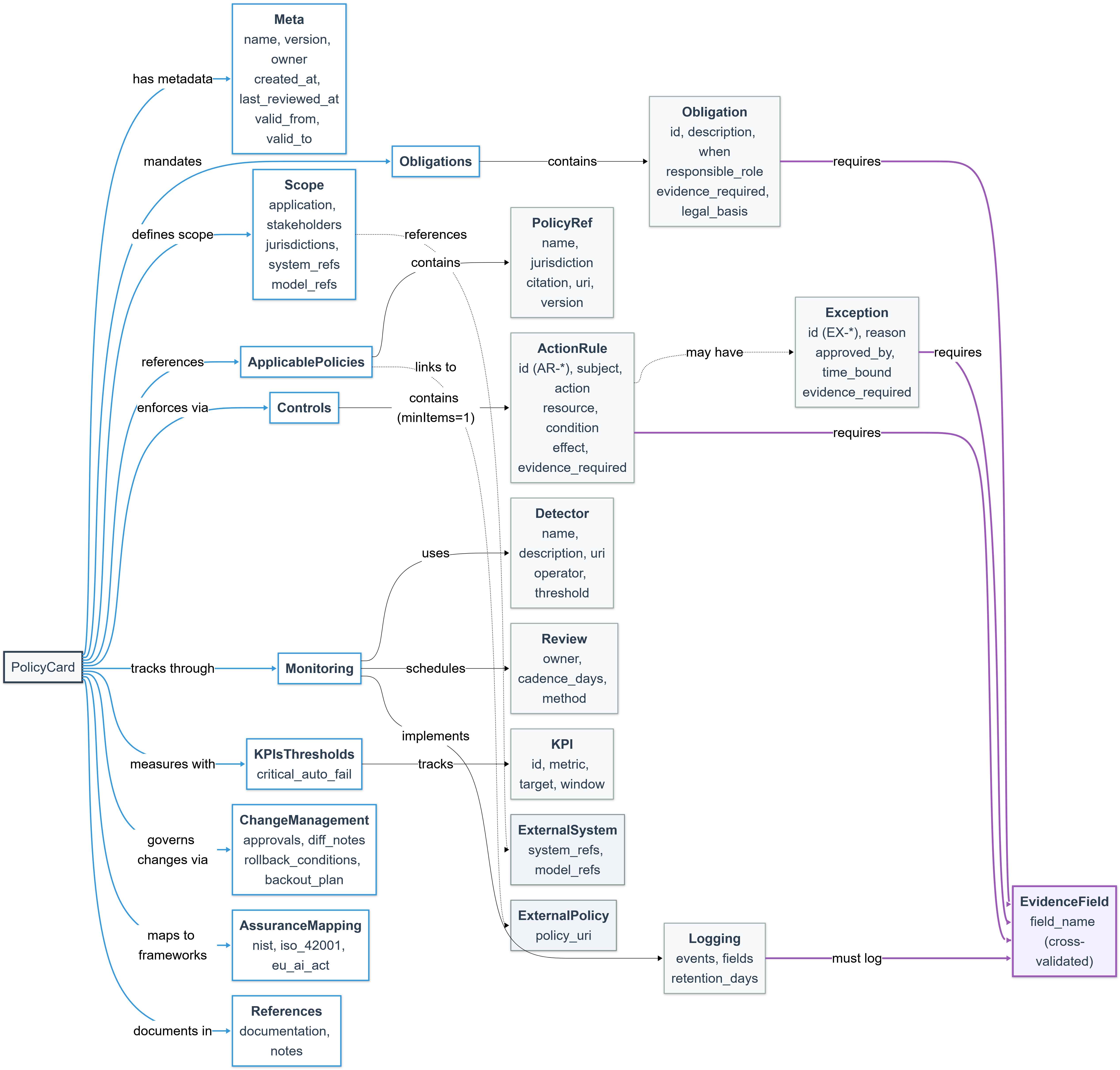}
\caption{Schema architecture. Sections and principal relationships.}
\label{fig:schema-diagram}
\end{figure}

The Policy Card schema is defined using JSON Schema 2020-12 \citep{jsonschema-core-2020-12}, providing a formally specified structure for encoding operational policies, obligations, and evidentiary controls (Figure \ref{fig:schema-diagram}). The schema serves as the normative backbone of the Policy Card framework. It defines mandatory fields, permitted data types, pattern constraints, and enumerations that govern how a valid card must be authored and validated. Its design goal is to enforce semantic discipline that reflects real compliance obligations. 

At the top level, the schema contains ten primary sections:
\begin{itemize}[]
    \item \texttt{meta}: Metadata about the Policy Card instance, including \texttt{name}, \texttt{version} (semantic versioning, \texttt{X.Y.Z}), \texttt{created\_at}, \texttt{last\_reviewed\_at}, and \texttt{valid\_from}/\texttt{valid\_to} timestamps, all compliant with RFC~3339 date-time formats.
    \item \texttt{scope}: Description of the deployment context, system boundaries, stakeholders, and jurisdiction. It defines the operational perimeter for which the card is valid.
    \item \texttt{applicable\_policies}: References to internal or external legal, regulatory, or organizational policies. Entries may include URIs, document identifiers, or citations.
    \item \texttt{controls}: Core operational rules represented as Attribute-Based Access Control (ABAC) tuples of $\langle$\textit{subject}, \textit{action}, \textit{resource}, \textit{condition}, \textit{effect}$\rangle$. The \texttt{effect} field is constrained to the values of \texttt{allow}, \texttt{deny}, or \texttt{require\_escalation}. This enables integration with enforcement engines and test harnesses. \texttt{exceptions} are explicit overrides to \texttt{controls}, each carrying an approver identifier, justification, and bounded validity period defined by a regex-validated pattern (\verb|^[0-9]+[mhdy]$|). These ensure traceable, time-limited deviations.
    \item \texttt{obligations}: Actions that must always occur under specified conditions, e.g., mandatory notifications, consent checks, safety statements. Obligations are auditable commitments linked to evidence fields.
    \item \texttt{monitoring}: Defines loggable events, data fields, detectors, and thresholds, together with minimum retention (\texttt{retention\_days >= 1}) and review cadence (\texttt{cadence\_days >= 1}). This section links operational telemetry with audit evidence.
    \item \texttt{kpis\_thresholds}: Quantitative performance and safety metrics (Key Performance Indicators, or KPIs) with \textit{critical\_auto\_fail} conditions. Violations of these thresholds trigger audit or escalation workflows.
    \item \texttt{change\_management}: Rules governing modification and review of the policy itself, including version control hooks and sign-off requirements.
    \item \texttt{assurance\_mapping}: Registry of tokens that link schema fields to assurance frameworks, e.g., NIST AI RMF, ISO/IEC 42001, EU AI Act Annex IV, Art. 72. Each token corresponds to a cross-referenced clause, allowing regulators and auditors to trace compliance programmatically.
    \item \texttt{references}: URIs and document identifiers for related artifacts such as Model Cards, Data Cards, System Cards, DPIAs, or risk assessments. This section ensures traceability and context for the policy definitions.
\end{itemize}
This structure enforces explicitness, completeness, and verifiability. Each section contributes a distinct dimension of assurance. It defines what is controlled, how it is monitored, and which standards it fulfills. This ensures that the card can serve as both documentation and enforcement artifact.

\subsection{Validation and Linting Rules}

A dedicated validator accompanies the schema to ensure consistency and correctness beyond structural validation. It uses a standard JSON Schema validator, e.g., Ajv 8.x, combined with custom linting logic that enforces semantic relationships between fields. The key validation checks include:

\begin{itemize}
    \item \emph{Schema conformance}: Ensures all required top-level keys are present and that no undefined properties are introduced (\texttt{additionalProperties: false}).
    \item \emph{Identifier and date validation}: Regular expressions and format constraints enforce patterns such as \texttt{AR-[A-Z0-9-]+} for action rules and RFC~3339 timestamps for temporal fields.
    \item \emph{Critical completeness}: Requires at least one ~ \texttt{critical\_auto\_fail} ~ KPI and ~ one token in each \texttt{assurance\_mapping} category.
    \item \emph{Temporal logic}: Ensures that \texttt{valid\_to >= valid\_from}, and that retention periods exceed review cadence (\texttt{retention\_days >= cadence\_days}).
    \item \emph{Consistency between sections}: Cross-checks that evidence fields listed in \texttt{monitoring} correspond to those referenced under \texttt{obligations} or \texttt{controls}.
    \item \emph{Enumerated effects}: Validates that every rule in \texttt{controls} uses a permitted \texttt{effect} token (\texttt{allow}, \texttt{deny}, or \texttt{require\_escalation}).
\end{itemize}
These lint rules prevent logically inconsistent or incomplete policy definitions and enforce uniform structure across all domain exemplars.

\begin{lstlisting}[language=json,caption={Abridged schema excerpt (illustrative). Full schema and validator are available in the public repository and archived via Zenodo.},label={lst:policy-card-schema}]
{
  "$id": "https://example.org/policy-cards/policy-card.schema.json",
  "$schema": "https://json-schema.org/draft/2020-12/schema",
  "title": "AI Policy Card",
  "description": "Deployment-layer, normative, audit-oriented policy for a specific AI system/agent in a defined context/jurisdiction.",
  "type": "object",
  "additionalProperties": false,

  "required": [
    "meta","scope","applicable_policies","controls","obligations",
    "monitoring","kpis_thresholds","change_management","assurance_mapping","references"
  ],

  "$defs": {
    "semver": { "type": "string", "pattern": "^\\d+\\.\\d+\\.\\d+$" },
    "rfc3339_datetime": { "type": "string", "format": "date-time" },
    "effect": { "type": "string", "enum": ["allow","deny","require_escalation"] },
    "attribute_expr": { "type": "string" },

    "id_action_rule": { "type": "string", "pattern": "^AR-[A-Z0-9-]+$" },

    "action_rule": {
      "type": "object",
      "additionalProperties": false,
      "required": ["id","subject","action","resource","condition","effect"],
      "properties": {
        "id": { "$ref": "#/$defs/id_action_rule" },
        "subject": { "type": "string" },         // actor/role
        "action": { "type": "string" },          // attempted operation
        "resource": { "type": "string" },        // domain object
        "condition": { "$ref": "#/$defs/attribute_expr" },
        "effect": { "$ref": "#/$defs/effect" },
        "evidence_required": {
          "type": "array", "items": { "$ref": "#/$defs/evidence_field" }
        }
        // "exceptions": [ ... ]
      }
    }

    // ... other $defs (detector, kpi, etc.)
  },

  "properties": {
    "meta": {
      "type": "object",
      "additionalProperties": false,
      "required": ["name","version","owner","created_at"],
      "properties": {
        "name": { "type": "string" },
        "version": { "$ref": "#/$defs/semver" },
        "owner": { "type": "string" },
        "created_at": { "$ref": "#/$defs/rfc3339_datetime" }
        // "last_reviewed_at", "valid_from", "valid_to"
      }
    },

    "scope": {
      "type": "object",
      "additionalProperties": false,
      "required": ["application","stakeholders","jurisdictions"],
      "properties": {
        "application": { "type": "string" },
        "stakeholders": { "type": "array", "items": { "type": "string" } },
        "jurisdictions": { "type": "array", "items": { "type": "string" } }
        // "system_refs", "model_refs"
      }
    },

    "applicable_policies": {
      "type": "array",
      "items": { "$ref": "#/$defs/policy_ref" } 
    },

    "controls": {
      "type": "object",
      "additionalProperties": false,
      "required": ["action_rules"],
      "properties": {
        "action_rules": {
          "type": "array",
          "items": { "$ref": "#/$defs/action_rule" },
          "minItems": 1
        }
      }
    },

    "monitoring": {
      "type": "object",
      "additionalProperties": false,
      "required": ["logging","detectors","review"],
      "properties": {
        "logging": {
          "type": "object",
          "required": ["events","fields","retention_days"]
          // details abridged
        },
        "detectors": { "type": "array" }, // abridged detector def
        "review": {
          "type": "object",
          "required": ["owner","cadence_days","method"]
        }
      }
    },

    "kpis_thresholds": {
      "type": "object",
      "required": ["kpis","critical_auto_fail"]
      // details abridged
    },

    "change_management": { "type": "object" },   // approvals, backout_plan...
    "assurance_mapping": { "type": "object" },   // nist, iso_42001, eu_ai_act
    "references": { "type": "object" }           // documentation, notes
  }
}
\end{lstlisting}

\subsection{Continuous Integration and Enforcement}

Because Policy Cards are code-validatable artifacts, they can be integrated directly into software development and deployment workflows. The validator can be executed as part of a CI/CD pipeline, rejecting deployments that violate schema rules or lints. Combined with runtime monitoring, this ensures that declared obligations and critical thresholds are continuously enforced.

Cards can also feed into policy gateways or agent orchestration frameworks, where enforcement engines evaluate \texttt{action\_rules} in real time and log outcomes defined by the \texttt{monitoring} section. This allows technical systems to demonstrate compliance through automated evidence capture rather than retrospective reporting.

By combining structural rigor with executable validation, the Policy Card schema and validator provide the necessary infrastructure for scalable, verifiable deployment-layer governance.

\section{Domain Exemplars}

The Policy Card framework was validated through multiple applied scenarios across sectors with distinct regulatory and evidentiary demands. This section presents two complete exemplars, one in retail banking and one in clinical triage, followed by a note on a defence-oriented case study. Each example demonstrates how the same schema structure can encode operational rules, evidence requirements, and assurance mappings specific to the domain, while remaining validator-clean and semantically consistent. The accompanying public repository includes scenarios demonstrating the application of the Policy Cards mentioned here.

\subsection{Retail Banking (Payments Agent)}

\subsubsection{Context and Scope}

Financial institutions operate within stringent regulatory frameworks such as PSD2 (EU), AMLD (Anti-Money Laundering Directive), and jurisdictional Know-Your-Customer (KYC) mandates. Automated agents that initiate or process payments must operate within well-defined authorization and escalation boundaries. The Retail Banking Policy Card formalizes these obligations for a payments-assistance agent integrated into a banking platform.

The scope of this Policy Card covers all transactions initiated through conversational or API-driven interfaces that trigger Faster Payments (GB/FPS) workflows. The agent may verify account balances, initiate payments, and confirm transaction outcomes, but must always respect compliance thresholds defined in the card.

\subsubsection{Operational Rules and Effects}

The \texttt{controls} section of the card defines explicit action rules structured as ABAC tuples. Example entries are listed in Table \ref{tab:banking-rules}.

\begin{table}[h]
\centering
\small
\begin{tabularx}{\textwidth}{@{}p{1.7cm}p{0.8cm}p{2.5cm}p{1.2cm}Yl@{}}
\toprule
ID & Subject & Action & Resource & Condition & Effect \\
\midrule
\texttt{AR-PAY- ALLOW- LOWRISK} & agent & \texttt{initiate\_payment} & payment & \texttt{kyc\_status == 'PASS' AND risk\_score < 0.70 AND device\_trust\_level >= 0.5
AND sanctions\_hit == false AND amount <= 2000 AND first\_time\_payee == false} & \texttt{allow} \\
\texttt{AR-PAY- ESCALATE- HIGHRISK} & agent & \texttt{initiate\_payment} & payment & \texttt{risk\_score >= 0.70 OR (first\_time\_payee == true AND amount >= 1000) OR device\_trust\_level < 0.5 OR geo\_mismatch == true OR (beneficiary\_screening\_score >= 0.6 AND beneficiary\_screening\_score < 0.8)}
 & \texttt{require\_escalation} \\
\texttt{AR-PAY- DENY- REDLINES} & agent & \texttt{initiate\_payment} & payment & \texttt{kyc\_status != 'PASS' OR sanctions\_hit == true OR mule\_account\_flag == true OR tamper\_check\_failed == true}
 & \texttt{deny} \\
\bottomrule
\end{tabularx}
\caption{Selected action rules from the Retail Banking exemplar.}
\label{tab:banking-rules}
\end{table}

Each ~~rule ~~links ~~directly ~to ~evidence ~~fields ~~enumerated ~~in ~the ~card, e.g., ~~~\texttt{kyc\_status}, ~~\texttt{risk\_score},~~~~~ \texttt{device\_trust\_level}, \texttt{beneficiary\_screening\_score}, \texttt{escalation\_id}, ensuring that every decision is auditable.

\subsubsection{Monitoring and Detectors}

The card mandates active risk detectors with numeric thresholds and declares all required log fields. Detectors include:
\begin{itemize}
    \item HighRiskScore (\texttt{>= 0.7})
    \item VelocitySpike (\texttt{> 3 within 1h}),
    \item GeoMismatch (\texttt{== 1}),
    \item SanctionsSoftMatchLowerBound (\texttt{>= 0.6}) and SanctionsSoftMatchUpperBound (\texttt{< 0.8}).
\end{itemize}
Logging retains fields such as \texttt{action\_rule\_id}, \texttt{amount}, \texttt{beneficiary\_screening\_score}, \texttt{device\_trust\_level}, \texttt{geo\_mismatch}, \texttt{kyc\_status}, \texttt{velocity\_1h}, and escalation references for 90 days. A 14-day review cadence is specified for detector performance and escalation analysis.

\subsubsection{KPIs and Critical Auto-Fail Conditions}

The \texttt{kpis\_thresholds} section defines:
\begin{itemize}
    \item \texttt{KPI-VIOL-RATE} (\texttt{violation\_rate}, the fraction of payment attempts whose final behavior violated the declared policy): target 0.01 over a 30-day window, and
    \item \texttt{KPI-ESC-SLA} (\texttt{escalation\_sla\_pct}, the percentage of escalations handled within the defined Service Level Agreement time window): target 95 over a 30-day window.
\end{itemize}
The ~~~\texttt{critical\_auto\_fail} ~~list ~~~includes: ~~~\texttt{transfer\_without\_KYC}, ~~~\texttt{transfer\_to\_sanctioned\_party},~ ~~~~~~and \texttt{override\_without\_two\_person\_approval}. Any breach triggers immediate review per the card's change-management backout plan.

\subsubsection{Obligations and Assurance Mapping}

Obligations ~~include ~user ~ notices ~ (\texttt{OB-USER-NOTICE-FEES-LIMITS}), ~ explicit ~ consent ~ recording ~ (\texttt{OB-RECORD-CONSENT}),~ escalation protocol enforcement (\texttt{OB-ESCALATION-PROTOCOL}), and audit-trail maintenance (\texttt{OB-AUDIT-TRAIL}). Assurance mappings use the tokens present in the card, e.g., \texttt{nist: GOVERN-1, MAP-1, MEASURE-1, MANAGE-1, MANAGE-3}; \texttt{iso\_42001: ISO42001-4, -8, -9, -10}; \texttt{eu\_ai\_act: EUAA-AnnexIV-3, -4, -6, -9, EUAA-Art72}, to support programmatic traceability. Specific, domain-related regulatory frameworks could be used.

\subsection{Clinical Triage (Sandbox)}

\subsubsection{Context and Scope}

Healthcare applications require strict adherence to clinical safety protocols and patient data protection rules such as the UK Data Protection Act (2018), the GDPR, and medical device regulations under MDR (EU 2017/745). The Clinical Triage Policy Card governs a triage assistant used to suggest risk categories and escalation paths based on user-provided symptoms. This system is explicitly non-diagnostic and must always defer to human judgment for high-risk or ambiguous cases. The Policy Card formalizes three key obligations:
\begin{enumerate}
    \item Provide only advisory outputs supported by evidence and disclaimers.
    \item Escalate any red-flag condition to qualified personnel or emergency services.
    \item Deny any autonomous diagnosis or prescription.
\end{enumerate}

\subsubsection{Operational Rules and Effects}

Rules defined under \texttt{controls} translate these obligations into machine-verifiable conditions. Example entries are listed in Table \ref{tab:triage-rules}.

\begin{table}[h]
\centering
\small
\begin{tabularx}{\textwidth}{@{}p{1.7cm}p{0.8cm}p{2.5cm}p{3.0cm}Yl@{}}
\toprule
ID & Subject & Action & Resource & Condition & Effect \\
\midrule
\texttt{AR-TRIAGE- ALLOW- ROUTINE} & agent & \texttt{propose\_triage} & \texttt{triage\_recommendation} & \texttt{vitals\_present == true AND risk\_level < 0.6 AND confidence\_score >= 0.8 AND red\_flag == false} & \texttt{allow} \\
\texttt{AR-TRIAGE- ESCALATE- REDFLAG} & agent & \texttt{propose\_triage} & \texttt{triage\_recommendation} & \texttt{red\_flag == true OR confidence\_score < 0.8 OR missing\_vitals\_count > 0 OR (patient\_age >= 65 AND symptom\_chest\_pain == true)} & \texttt{require\_escalation} \\
\texttt{AR-TRIAGE- DENY- AUTONOMY} & agent & \texttt{issue\_diagnosis} & \texttt{diagnosis\_statement} & \texttt{true} & \texttt{deny} \\
\texttt{AR-TRIAGE- DENY- PRESCRIPTION} & agent & \texttt{issue\_prescription} & \texttt{medication\_order} & \texttt{true} & \texttt{deny} \\
\bottomrule
\end{tabularx}
\caption{Selected action rules from the Clinical Triage exemplar.}
\label{tab:triage-rules}
\end{table}

An exception is defined for remote clinical sites (\texttt{EX-TRIAGE-REMOTE-CLINIC-OVERRIDE}), time-bound to 12h, requiring \texttt{two\_person\_approval\_id} and a \texttt{callback\_reference} as evidence.

\subsubsection{Monitoring and Detectors}

The ~~\texttt{monitoring} ~~section ~~specifies ~~logging ~~of ~~all ~~intake ~~and ~~triage events, ~~with fields including \texttt{action\_rule\_id}, \texttt{confidence\_score}, ~~~~\texttt{missing\_vitals\_count},~ ~~~\texttt{patient\_age}, ~~~\texttt{triage\_category}, ~~~~\texttt{model\_version}, ~~~~~~and \texttt{detector\_snapshots\_uri}. Retention is 365 days. Detectors include LowConfidence (\texttt{< 0.8}), MissingVitals (\texttt{> 0}), BiasDrift (\texttt{>= 0.1}), RedFlagDetected (\texttt{== 1}), and IntakeTimeout (\texttt{> 300} seconds). Review cadence is 14 days under the clinical governance board.

\subsubsection{KPIs and Critical Auto-Fail Conditions}

The \texttt{kpis\_thresholds} section defines operational metrics:
\begin{itemize}
    \item \texttt{KPI-HANDOVER-P50} (median handover latency, i.e., the median time it takes, in minutes, for a triage case to be handed over to a clinician and reviewed): target 15 minutes over 30 days, and
    \item \texttt{KPI-ESC-PRECISION} (escalation precision, i.e., the fraction of triggered escalations that clinicians later judge as warranted): target 0.9 over 30 days.
\end{itemize}
Each KPI provides quantitative assurance of safety and responsiveness. \texttt{critical\_auto\_fail} events include: \texttt{autonomous\_diagnosis\_emitted}, \texttt{autonomous\_prescription\_emitted}, and \texttt{triage\_without\_vitals}. Violations trigger the backout plan specified in change-management.

\subsubsection{Obligations and Assurance Mapping}

Obligations ~~~include ~~~mandatory ~~~clinician ~~review ~~(\texttt{OB-CLINICIAN-REVIEW}), ~~~~patient ~~notices~~~ (\texttt{OB-PATIENT-NOTICE}), and model/version traceability (\texttt{OB-MODEL-VERSION-TRACE}). The assurance mapping uses the tokens enumerated in the card, e.g., \texttt{nist}, \texttt{iso\_42001}, and \texttt{eu\_ai\_act} keys, to support auditable alignment with external frameworks. Specific, domain-related frameworks could be used.

\subsection{Defence Mission-Planning}

A separate Policy Card was created for a defence mission-planning assistant supporting coalition UAV coordination. This card demonstrates the applicability of Policy Cards in safety-critical and classified contexts. Its primary rules include: prohibiting autonomous kinetic actions (\texttt{deny}), requiring escalation for target designations (\texttt{require\_escalation}), enforcing blue-force deconfliction, and logging immutable override audits. The example validates that the schema accommodates multi-jurisdictional Rules of Engagement (ROE) and security-classified use cases without structural modification.

\subsection{Discussion of Domain Generality}

Across these exemplars, the Policy Card schema consistently captures, (1), explicit, testable control logic, (2), traceable evidence bindings, and, (3), verifiable links to recognized assurance frameworks. The validator ensures that each domain-specific configuration remains structurally and semantically compliant while encoding obligations unique to its regulatory environment. The result is a portable, extensible mechanism for expressing governance-by-construction, demonstrating that deployment-layer policy can be standardized without sacrificing domain specificity.

\section{Integration: Declare-Do-Audit \& Stress Testing}

\subsection{Overview}

The Policy Card framework operationalizes compliance through an auditable lifecycle that aligns policy declaration, system execution, and continuous verification. This lifecycle (Declare, Do, Audit) ensures that AI deployments can demonstrate conformance both at design time and during live operation. Each stage defines clear control points where Policy Card content interacts with implementation logic, telemetry systems, and assurance workflows.

\begin{figure}[h]
\centering
\includegraphics[width=0.50\linewidth]{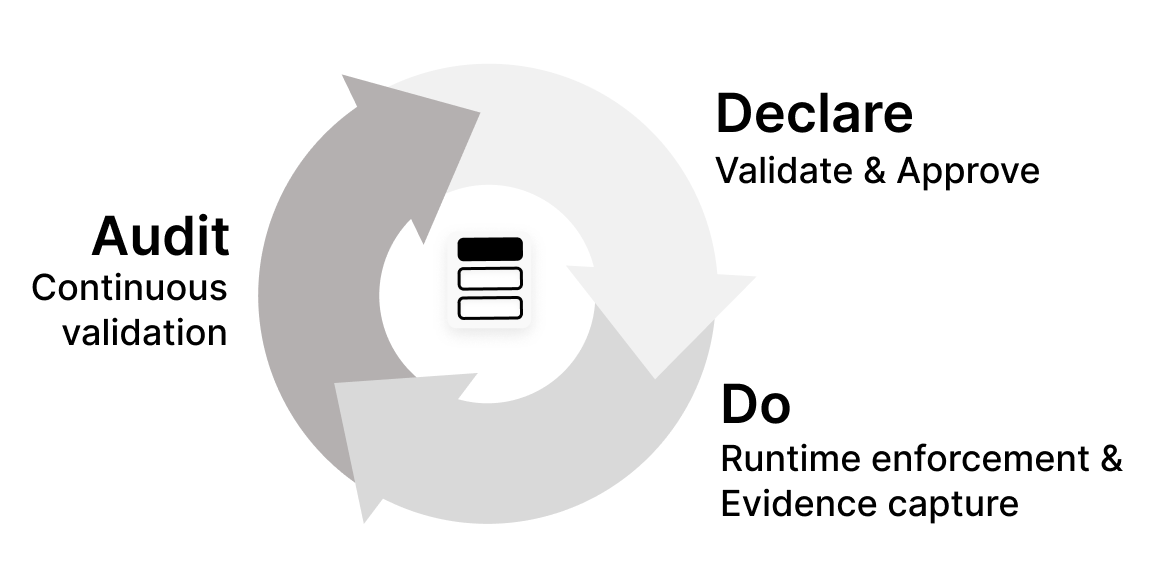}
\caption{The Policy Card sits at the heart of the governance lifecycle. It supports policy declaration, controlled execution, evidence capture and automated audit feedback.}
\label{fig:declare-do-audit}
\end{figure}

The integration pattern treats the Policy Card as the authoritative specification of deployment rules (Figure \ref{fig:declare-do-audit}). During declaration, it binds policy parameters to the system configuration and records baseline evidence. During execution, it governs runtime decisions and logging. During audit, it supports verification of metrics, thresholds, and exception-handling outcomes.

\subsection{Declare Phase: Binding and Verification}

The \emph{Declare} phase occurs before deployment. Here, the Policy Card is registered within the deployment pipeline and validated using the schema validator described in Section \ref{sec:schema}. The validator enforces syntactic correctness, semantic integrity, and cross-section consistency, ensuring the card is fit for operational use.

Once validated, the Policy Card's version identifier, cryptographic hash, and applicable context are stored in a configuration registry, e.g., Git-based version control or a model registry such as MLflow. The declaration process establishes baseline evidence:
\begin{itemize}
    \item Policy version and checksum.
    \item Declared critical thresholds, e.g., KPI violation-rate targets.
    \item Approved exceptions and their expiration timestamps.
    \item Crosswalk tokens binding policy clauses to external frameworks.
\end{itemize}

In regulated environments, this declaration step can be formally approved by internal compliance officers or auditors. 
Policy Cards enable a clean separation of responsibilities. Model engineers and agent deployers are not required to interpret regulatory obligations directly. Their task is simply to ensure that the deployed system conforms to its assigned Policy Card, while compliance and governance specialists define the applicable rules, legal requirements, and assurance criteria within the card itself. The declarative process prevents policy drift by ensuring that any modification triggers a new version requiring re-approval.

\subsection{Do Phase: Execution and Evidence Capture}

During the \emph{Do} phase, the deployed AI system or agent operates under the governance of the active Policy Card. The card is attached to the agent/system. Each system component interprets the card's \texttt{controls} and \texttt{obligations} sections to decide whether actions are allowed, denied, or require escalation. Enforcement occurs via integrated policy gateways or API-level middleware.

For instance, in the Retail Banking exemplar, an attempted payment above the risk threshold (\texttt{risk\_score >= 0.70}) triggers the escalation rule \texttt{AR-PAY-ESCALATE-HIGHRISK}, requiring supervisory approval before execution.~ ~In ~the ~Clinical ~Triage ~~exemplar, ~a ~~red-flag symptom ~or ~missing ~vitals ~triggers ~~~~\texttt{AR-TRIAGE-ESCALATE-REDFLAG}, generating a review task for clinical staff. All such decisions emit structured evidence defined in the \texttt{monitoring} section: timestamps, rule IDs, conditions, detector snapshots, and outcome tags.

The runtime system continuously emits structured logs into a secure evidence store. Detectors run asynchronously, verifying adherence to retention and review cadences. The monitoring subsystem enforces auto-fail triggers defined in the Policy Card, e.g., \texttt{transfer\_without\_KYC} or \texttt{triage\_without\_vitals}, and flags violations in real time. This creates an immutable, verifiable record of all operational decisions.

\subsection{Audit Phase: Automated and Continuous Assurance}

The \emph{Audit} phase consolidates all evidence emitted during execution and tests it against the declared policy. Two complementary modes are supported:
\begin{enumerate}
    \item Automated CI-based audits, performed in continuous integration pipelines after each deployment iteration. The validator replays logged events, comparing evidence against declared thresholds, obligations, and assurance mappings.
    \item Post-market or continuous audits, executed by external auditors or internal assurance teams. These involve replaying historical data to verify compliance trajectories over time.
\end{enumerate}

The auditing system cross-references evidence with Policy Card identifiers, ensuring that all required fields, detector outputs, and KPI statistics meet or exceed targets. Violations automatically trigger incident reports defined in the \texttt{change\_management} section. This ensures that deviations from declared policies are captured and resolved systematically.

\subsection{Stress Testing and Simulated Assurance}

Policy Cards also enable \emph{stress testing} of governance constraints. By using synthetic or simulated data, auditors and developers can test how systems behave under edge conditions, e.g., fraudulent transaction bursts, missing clinical data, or latency spikes. Because the Policy Card defines both allowed actions and failure thresholds, these stress tests provide a quantitative method to evaluate the robustness of enforcement mechanisms.

In practice, stress tests can be implemented by executing the AI system against controlled test harnesses configured to read the same Policy Card instance as the live system. The harness injects synthetic inputs and verifies whether the enforcement engine, e.g., the agent, correctly applies each \texttt{allow}, \texttt{deny}, or \texttt{require\_escalation} rule and whether evidence capture meets defined retention and completeness requirements. Results are recorded as structured compliance reports, supporting certification and regulatory audits.

\subsection{Integration Benefits}

The Declare-Do-Audit cycle transforms the Policy Card from a static documentation artifact into a live governance interface. It ensures that declared policies are:
\begin{itemize}
    \item \emph{Executable}: directly enforceable by system components, e.g., the AI agents themselves.
    \item \emph{Measurable}: equipped with KPIs and detectors for evidence-based assessment.
    \item \emph{Auditable}: designed for automated replay and continuous validation.
\end{itemize}

This integration pattern unifies policy definition, operational behavior, and verification. The agent enforces the card, initiates escalations, emits signed evidence, and can self-test under simulation using the same Policy Card. Together with stress testing, this yields continuous assurance of compliance and resilience under real-world and adversarial conditions.

\section{Forward-Looking Features}

\subsection{Agent-Readable Policies and Multi-Agent Governance}

Since Policy Cards are machine-interpretable and semantically explicit, they enable AI agents to ingest, reason about, and enforce their own operational policies. An agent can parse the card, evaluate the scope of its permitted actions, and determine escalation or reporting duties without human intervention.

An agent operating under a Policy Card can perform three autonomous checks before any action:
\begin{enumerate}
    \item \emph{Authorization evaluation}: Verifying that the proposed action satisfies all preconditions defined in the \texttt{controls} section.
    \item \emph{Obligation assessment}: Confirming that all mandatory evidence requirements (from \texttt{obligations}) can be satisfied prior to execution.
    \item \emph{Contextual validation}: Checking environmental constraints such as jurisdiction, operational mode, or time window against the \texttt{scope} section.
\end{enumerate}
This design allows \emph{policy self-awareness}. The agent continuously reconciles its behavior against its declared policy. When implemented within multi-agent systems, this principle scales into a distributed assurance mesh. Each agent carries its own Policy Card, publishes verifiable compliance states, and interacts with others under mutual constraints (Figure \ref{fig:multi-agent-systems}).

In a collaborative scenario, such as the additional multi-agent Clinical Trial Coordinator exemplar provided in the codebase, agents could exchange digital attestations representing partial fulfillment of obligations, e.g., data anonymization, consent verification, statistical integrity. These attestations could be verified by peers and aggregated into a composite assurance graph. This can enable federated compliance, where trust does not rely on a single centralized auditor but emerges from a network of mutually validating agents.

Such agent-readable governance is an essential prerequisite for scalable oversight of autonomous ecosystems. It transforms compliance from a static process into a dynamic, computational function embedded directly within multi-agent architectures.

\subsection{Security and Cryptographic Extensions}

\begin{figure}[h]
\centering
\includegraphics[width=0.45\linewidth]{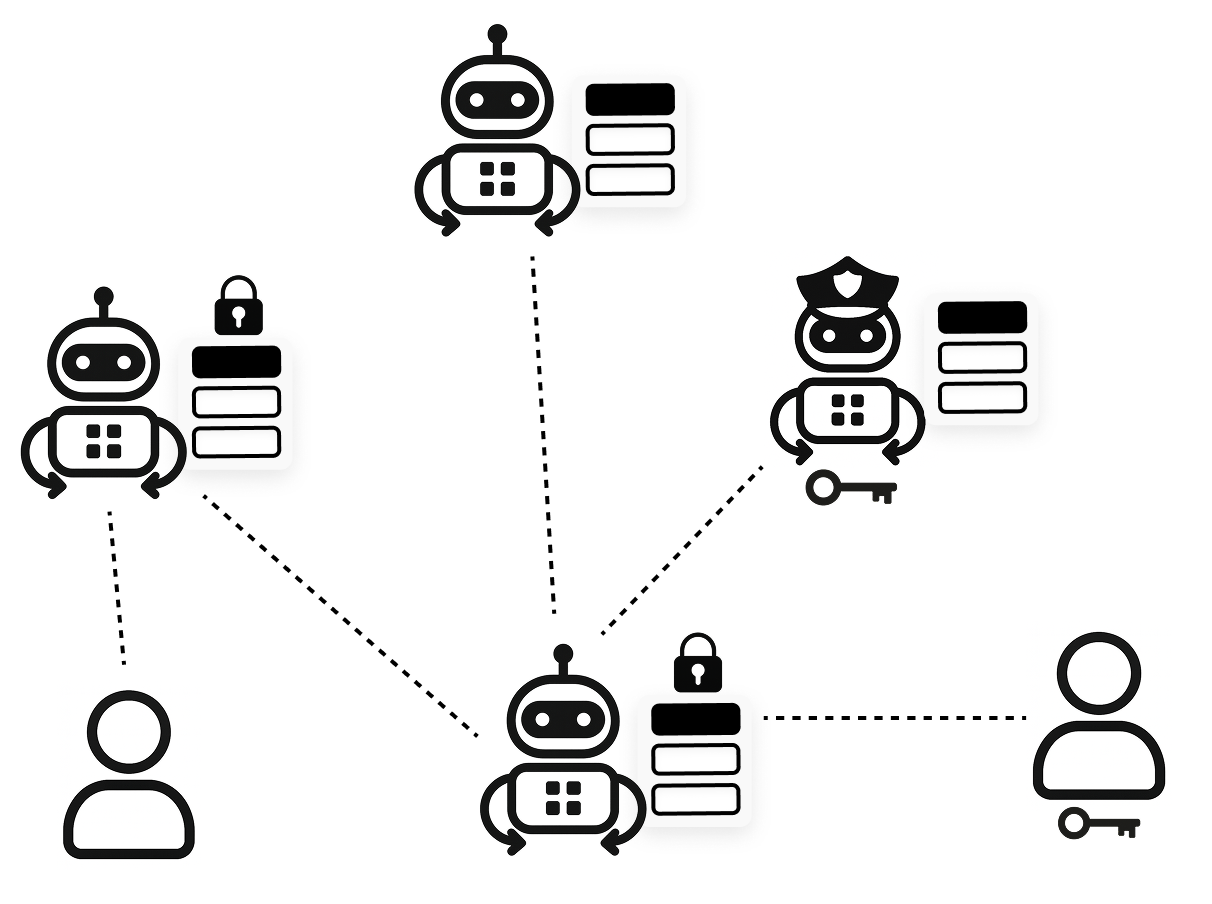}
\caption{In multi-agent systems each agent carries its own Policy Card, building a distributed governance mesh. A Policy Card may be encrypted and parts of it visible only to oversight agents or regulators. Agents can produce zero-knowledge proofs asserting compliance with certain policy conditions without revealing internal data.}
\label{fig:multi-agent-systems}
\end{figure}

Future implementations of Policy Cards can integrate cryptographic techniques to provide verifiable, privacy-preserving assurance. Two mechanisms are particularly relevant:
\begin{enumerate}
    \item \emph{Key-Scoped Policy Encryption.} A Policy Card may be encrypted using public-key infrastructure (PKI), where only authorized agents, identified by their signing keys, can decrypt and interpret the card. This ensures that operational policies remain confidential while still being enforceable. Oversight agents or regulators can hold verification keys that allow them to confirm adherence without exposing proprietary rules.
    \item \emph{Zero-Knowledge Proofs for Policy Adherence.} Agents can produce zero-knowledge proofs (ZKPs) asserting that they have complied with certain policy conditions, e.g., no unauthorized data exfiltration, no unverified transaction execution, without disclosing the underlying data. The Policy Card defines which metrics are provable and provides cryptographic templates for proof construction. These proofs can be logged as part of the \texttt{evidence\_required} or \texttt{assurance\_mapping} sections, allowing auditors to verify compliance without direct data access.
\end{enumerate}
When combined, encryption and ZKPs create a confidential assurance layer. Operational compliance can be verified across distributed environments without revealing sensitive information (Figure \ref{fig:multi-agent-systems}). This is particularly useful in cross-border, regulated, or classified deployments where data-sharing restrictions apply. In such cases, Policy Cards act as structured metadata describing both the operational policy and its cryptographic verification envelope.

\subsection{Ethical Policy Composition}

Beyond regulatory and operational constraints, Policy Cards can also encode ethical policies. These can be understood as structured commitments reflecting societal or organizational values. While ethics are often expressed as narrative guidelines, Policy Cards enable their translation into formal, executable constructs.

Ethical obligations can be represented as specialized entries within the `obligations` or `controls` sections. For example, a card may include rules preventing discriminatory outcomes, e.g., fairness thresholds, obligations to provide transparency notices, or prohibitions against manipulative behavior in user interactions. These constraints can be quantified, logged, and audited like any other operational policy.

Conflicts between ethical obligations and performance objectives can be resolved through priority ordering. The schema allows for explicit precedence definitions. For instance, an obligation could be tagged \texttt{ethical\_priority: high} to override conflicting optimization objectives. This mechanism provides a formal way to resolve competing ethical principles at runtime while maintaining traceability for audit.

Embedding ethical constraints in Policy Cards advances the notion of operationalized ethics, ethical principles implemented as enforceable, measurable, and machine-verifiable rules. This approach ensures that normative commitments are not aspirational but are directly encoded into system behavior and governance pipelines.

\subsection{Synthesis}

The forward-looking capabilities outlined here; agent-readable governance, cryptographic assurance, and ethical policy composition, extend Policy Cards beyond compliance documentation. They position the framework as a foundation for autonomous assurance ecosystems, where policies are computational, verifiable, and privacy-preserving. In such a system, both human regulators and autonomous agents participate in maintaining conformance, establishing a path toward distributed, scalable, and trustworthy AI governance.

\section{Standards Crosswalk \& Regulatory Alignment}

This section summarizes how Policy Cards align with three major governance and assurance frameworks; the NIST AI Risk Management Framework (1.0) \citep{nist}, the ISO/IEC 42001:2023 (AI Management System Standard) \citep{iso_IEC}, and the EU AI Act (Annex IV / Article 72) \citep{european_parliament_and_council_regulation_2024}. The mappings below are maintained in the repository, ensuring consistency with the \texttt{assurance\_mapping} tokens embedded in all validated Policy Cards. The purpose of this alignment is to demonstrate that Policy Cards can serve as an interoperable assurance layer, connecting technical system configuration with high-level governance and compliance frameworks. The same methodology can be extended to other national or industry-specific standards as needed.

\begin{figure}[h]
\centering
\includegraphics[width=\textwidth]{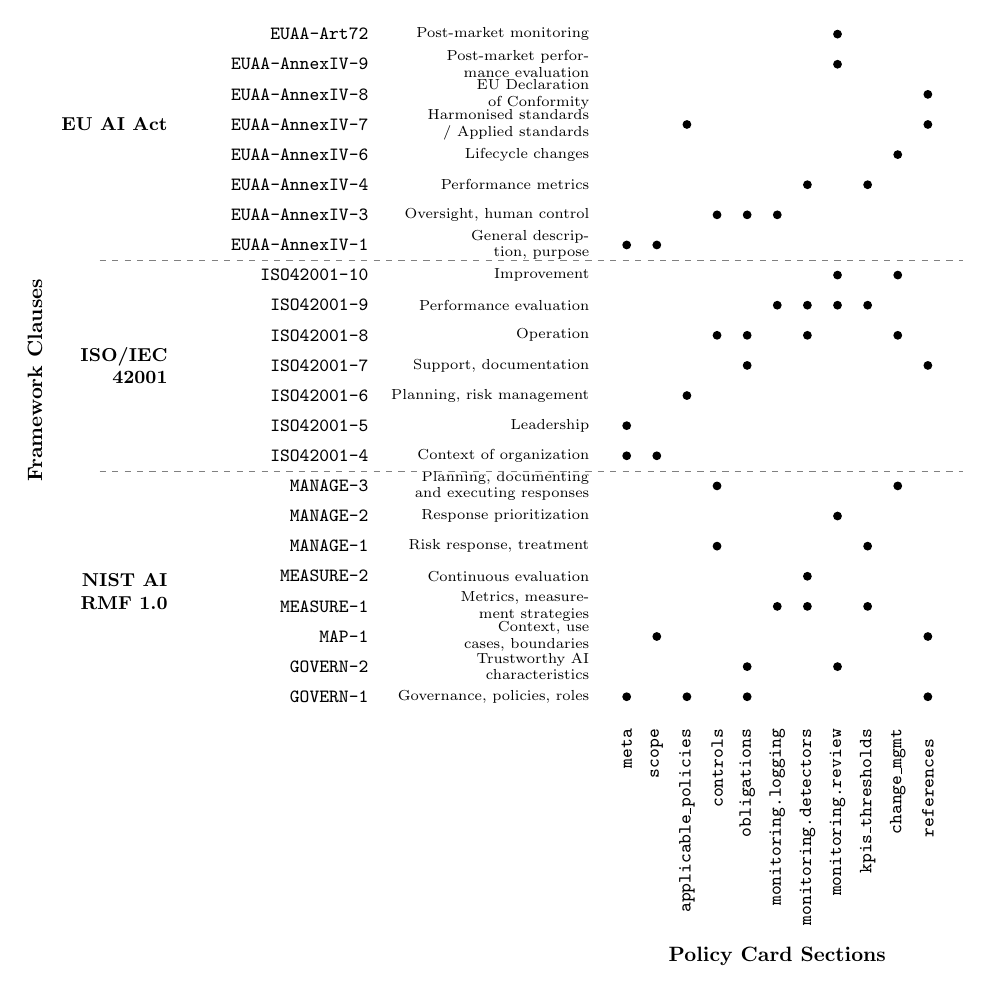}
\caption{Policy Card coverage across NIST AI RMF 1.0, ISO/IEC 42001, and the EU AI Act. Filled dots indicate mappings between Policy Card sections and framework clauses, demonstrating interoperability and completeness of the Policy Card governance layer.}
\label{fig:crosswalk-matrix}
\end{figure}

\paragraph{Crosswalk Overview.} Each Policy Card section has a normative correspondence to clauses and functions within the three frameworks, as summarized in Table \ref{tab:standards-crosswalk} and Figure \ref{fig:crosswalk-matrix}. The crosswalk enables automated validation of completeness. 

\paragraph{Token Registry.} Policy Cards use canonical tokens as lightweight abstractions for framework clauses. For example:
\begin{itemize}
    \item NIST AI RMF 1.0: \texttt{GOVERN}, \texttt{MAP}, \texttt{MEASURE}, \texttt{MANAGE} (core functions).
    \item ISO/IEC 42001: \texttt{ISO42001-4} to \texttt{ISO42001-10} (Context to Improvement).
    \item EU AI Act: \texttt{EUAA-AnnexIV-1} to \texttt{EUAA-AnnexIV-9} and \texttt{EUAA-Art72} (technical documentation and post-market monitoring).
\end{itemize}

\begin{table}[h]
\centering
\small
\begin{tabularx}{\textwidth}{@{}p{3.2cm}Xp{2.2cm}p{2.3cm}p{2.7cm}@{}}
\toprule
Policy Card Section & Purpose / Role & NIST AI RMF Tokens & ISO/IEC 42001 Tokens & EU AI Act Tokens \\
\midrule
\texttt{meta} & Captures metadata, ownership, and governance context. & \texttt{GOVERN-1} & \texttt{ISO42001-5}, \texttt{-4} & \texttt{EUAA-AnnexIV-1} \\
\texttt{scope} & Defines use context, boundaries, and jurisdictions. & \texttt{MAP-1} & \texttt{ISO42001-4} & \texttt{EUAA-AnnexIV-1} \\
\texttt{applicable\_policies} & References applicable laws, standards, and internal policies. & \texttt{GOVERN-1} & \texttt{ISO42001-6} & \texttt{EUAA-AnnexIV-7} \\
\texttt{controls.action\_rules} & Expresses enforceable operational control logic. & \texttt{MANAGE-1}, \texttt{-3} & \texttt{ISO42001-8} & \texttt{EUAA-AnnexIV-3} \\
\texttt{obligations} & Embeds mandatory behaviors such as notice and consent. & \texttt{GOVERN-2}, \texttt{-1} & \texttt{ISO42001-8}, \texttt{-7} & \texttt{EUAA-AnnexIV-3} \\
\texttt{monitoring.logging} & Defines telemetry and evidence retention for audit. & \texttt{MEASURE-1} & \texttt{ISO42001-9} & \texttt{EUAA-AnnexIV-3} \\
\texttt{monitoring.detectors} & Defines automated performance and anomaly detection. & \texttt{MEASURE-1}, \texttt{-2} & \texttt{ISO42001-9}, \texttt{-8} & \texttt{EUAA-AnnexIV-4} \\
\texttt{monitoring.review} & Ensures post-market monitoring and review cadences. & \texttt{GOVERN-2}, \texttt{MANAGE-2} & \texttt{ISO42001-10}, \texttt{-9} & \texttt{EUAA-Art72}, \texttt{EUAA-AnnexIV-9} \\
\texttt{kpis\_thresholds} & Specifies measurable assurance metrics and red lines. & \texttt{MEASURE-1}, \texttt{MANAGE-1} & \texttt{ISO42001-9} & \texttt{EUAA-AnnexIV-4} \\
\texttt{change\_management} & Controls lifecycle updates, approval, and rollback. & \texttt{MANAGE-3} & \texttt{ISO42001-10}, \texttt{-8} & \texttt{EUAA-AnnexIV-6} \\
\texttt{references} & Establishes document traceability and applied standards. & \texttt{GOVERN-1}, \texttt{MAP-1} & \texttt{ISO42001-7} & \texttt{EUAA-AnnexIV-7}, \texttt{-8} \\
\bottomrule
\end{tabularx}
\caption{Standards crosswalk: Mapping Policy Card sections to NIST AI RMF, ISO/IEC 42001, and EU AI Act tokens.}
\label{tab:standards-crosswalk}
\end{table}

A typical \texttt{assurance\_mapping} entry within a Policy Card is shown in Listing \ref{lst:assurance-mapping}.
\begin{lstlisting}[language=json,caption={Example assurance mapping structure.},label={lst:assurance-mapping}]
"assurance_mapping": {
  "nist": ["GOVERN-1", "MAP-1", "MEASURE-1", "MANAGE-1"],
  "iso_42001": ["ISO42001-4", "ISO42001-8", "ISO42001-9"],
  "eu_ai_act": ["EUAA-AnnexIV-3", "EUAA-AnnexIV-4", "EUAA-Art72"]
}
\end{lstlisting}

\paragraph{Regulatory Readiness.} A Policy Card can be programmatically checked to ensure that all governance, mapping, measurement, and management requirements within a given regulatory framework are represented. In addition, by embedding crosswalk tokens, Policy Cards operationalize different assurance frameworks simultaneously. The approach converts abstract regulatory requirements into verifiable configuration artifacts. This design also makes Policy Cards suitable for adoption as a harmonization layer between national assurance regimes, enabling consistent compliance reporting across jurisdictions.

\section{Conclusion \& Future Work}

This work introduces Policy Cards as a machine-readable, runtime-layer artifact that unifies operational rules, obligations, evidence requirements, and regulatory mappings into a single, auditable specification. Across all exemplars and schema components, the framework demonstrates that compliance can be expressed as executable policy used by both humans and AI agents. By formalizing what actions are allowed, what evidence is required, and how assurance is measured, Policy Cards transform AI governance into a verifiable engineering discipline.

The proposed schema, validator, and integration model show that regulatory expectations, such as those defined by NIST AI RMF, ISO/IEC 42001, and the EU AI Act, can be instantiated directly within the same structure used for runtime enforcement. The resulting linkage between Declare-Do-Audit phases provides a continuous, testable chain from policy intent to operational evidence. This mechanism connects abstract compliance frameworks with concrete system behavior.

We envision several research and engineering directions that emerge based on the Policy Cards framework:
\begin{enumerate}
    \item Formal semantics and enforcement back-ends. Translating Policy Card logic into executable policy languages, e.g., Rego (via OPA), Cedar, XACML/ALFA, or CEL, would integrate the framework with existing systems for enforcement and formal verification.
    \item Automated card synthesis. Tooling could generate baseline Policy Cards automatically from CI pipelines, model metadata, or existing governance registries, reducing authoring burden and ensuring completeness.
    \item Multi-jurisdictional composition. Modular Policy Cards could be layered to reflect overlapping legal and sector requirements, supporting global deployments with local overrides.
    \item Cryptographic attestation maturity. Further development of zero-knowledge proof and key-scoped encryption prototypes will be required to make confidential assurance practical at scale.
    \item Institutional adoption and legal acceptance. Engagement with regulators and standards bodies can establish Policy Cards as a reference format for technical documentation and conformity assessment for autonomous AI agents.
\end{enumerate}

In essence, Policy Cards establish a new operational language of trust between governance and practice. For regulators and oversight bodies, they provide concrete, machine-verifiable evidence that declared principles are enforced in real deployments. For practitioners, they turn governance from an external constraint into a trusted engineering tool, that improves clarity, accountability, and system resilience. In future autonomous ecosystems, both humans and AI agents will be able to rely on Policy Cards as guiding artifacts for coordination, assurance, and ethical direction.

\section{Availability \& Licensing}

All source artifacts described in this paper, including the Policy Card schema, validator, crosswalk mappings, and validated exemplars are available in the public repository at:

\url{https://github.com/symbiotic-dynamics/policy-cards}

The Policy Card JSON Schema is released under CC0 1.0 Universal Public Domain Dedication, allowing unrestricted reuse.
Code/tooling artifacts are released under the Apache License 2.0, permitting use, modification, and redistribution with preservation of LICENSE and NOTICE.
Docs/specs/templates/examples are released under CC BY 4.0, allowing sharing and adaptation with attribution.
A permanent archival copy of the repository, including schema versions and example Policy Cards, is maintained on Zenodo under \url{https://doi.org/10.5281/zenodo.17392820}.

The paper text and figures are released under CC BY 4.0, allowing sharing and adaptation with attribution.

\paragraph{Disclaimer.}
This work and the associated artifacts are provided \emph{as is}, without any warranty or guarantee of fitness for safety-critical or operational use. The author accepts no responsibility or liability for the use or misuse of the materials, which are intended solely for research and educational purposes.

\paragraph{Acknowledgements.} The author thanks all early reviewers who provided technical and conceptual feedback on the Policy Cards framework. In particular, the author acknowledges Dr.~Ryan Cunningham for detailed comments on the framework and his insights on cryptographic extensions.

\bibliographystyle{unsrtnat}
\bibliography{bibliography}
\end{document}